\documentclass[runningheads]{llncs}

\usepackage[utf8]{inputenc}
\usepackage{amsmath, amssymb, amsfonts, dsfont, mathrsfs, dirtytalk, tikz-cd, adjustbox, url, nicefrac, booktabs, array, graphicx,todonotes,longtable, tabu, bm, cite, todonotes, comment, multirow,tabularx,epsfig,parskip,subcaption}
\usepackage[font=footnotesize]{caption}
\usepackage[T1]{fontenc}
\usepackage[english]{babel}
\usepackage{hyperref}
\usepackage[capitalise]{cleveref}

\newcommand{\reals}{\mathbb{R}} 
\newcommand{\DKL}[2]{D_{\text{KL}}\left[#1\ \vert\vert\ #2 \right]}

\begin{document}
\title{On the Variational Costs of Changing Our Minds}

\author{David Hyland$^1$ and Mahault Albarracin$^2$}

\authorrunning{D. Hyland and M. Albarracin}
\institute{University of Oxford, Oxford, United Kingdom \and VERSES AI Research Lab, Los Angeles, CA 90016, USA
\email{david.hyland@cs.ox.ac.uk}\\
\email{mahault.albarracin@verses.ai}}
\maketitle
\begin{abstract}
The human mind is capable of extraordinary achievements, yet it often appears to work against itself. It actively defends its cherished beliefs even in the face of contradictory evidence, conveniently interprets information to conform to desired narratives, and selectively searches for or avoids information to suit its various purposes. Despite these behaviours deviating from common normative standards for belief updating, we argue that such `biases' are not inherently cognitive flaws, but rather an adaptive response to the significant pragmatic and cognitive costs associated with revising one's beliefs. This paper introduces a formal framework that aims to model the influence of these costs on our belief updating mechanisms.

We treat belief updating as a motivated variational decision, where agents weigh the perceived `utility' of a belief against the informational cost required to adopt a new belief state, quantified by the Kullback-Leibler divergence from the prior to the variational posterior. We perform computational experiments to demonstrate that simple instantiations of this resource-rational model can be used to qualitatively emulate commonplace human behaviours, including confirmation bias and attitude polarisation. In doing so, we suggest that this framework makes steps toward a more holistic account of the motivated Bayesian mechanics of belief change and provides practical insights for predicting, compensating for, and correcting deviations from desired belief updating processes.

\end{abstract}

``\emph{[H]uman reason is both biased and lazy. Biased because it overwhelmingly finds justifications and arguments that support the reasoner’s point of view, lazy because reason makes little effort to assess the quality of the justifications and arguments it produces.}'' (Mercier and Sperber, 2017, p. 9)~\cite{mercier2017enigma}.

\section{Introduction}

Humanity faces an increasingly paradoxical epistemic problem. Never before have people been able to obtain so much information so quickly, yet at the same time, many societies have become increasingly polarised and paralysed by conflicting narratives. A feature that is common to several manifestations of this predicament, including public health crises and conspiracy theorising, is the presence of actors who tenaciously defend beliefs long after the balance of evidence has shifted. What, exactly, makes changing our minds so hard?

A natural approach to answering this question may begin by supposing a normative standard or benchmark against which to compare the actual processes of human belief change. The primary normative model for rational belief updating is Bayesian reasoning. According to this model, probabilistic beliefs should be adjusted proportionally to the strength of evidence according to Bayes' rule. However, the persistent discrepancy between the Bayesian standard and actual human belief updating raises questions about whether our epistemic processes are inherently irrational or if something is missing from the traditional picture~\cite{mandelbaum2019troubles}. 

The Bayesian paradigm largely remains silent on \textit{why} humans fall short of its ideals, primarily due to its assumptions that belief-revision is cost-free and that the driver of epistemic processes should be probabilistic coherency~\cite{jones2011pinning}. In practice, revising one's beliefs incurs metabolic costs, cognitive effort, and crucially, pragmatic risks and opportunities. A scientist retracting a cherished hypothesis, a politician breaking ranks with their party, or a public figure admitting error each pay tangible costs that a cost‑free Bayesian calculus does not account for. Without a principled way to model the effect of such costs on belief revision, apparent ``irrationalities'' including confirmation bias~\cite{nickerson1998confirmation}, motivated reasoning~\cite{kunda1990case}, attitude polarisation~\cite{lord1979biased}, and belief persistence~\cite{ecker2022psychological} seem like fundamental flaws in human cognition.

We argue that such apparent deviations from Bayesian norms are adaptive responses of agents operating under motivational considerations and real resource constraints. In particular, we formalise cognitive belief-change costs using the KL divergence to quantify informational distances between belief states, representing the `informational work' required for belief state transitions. Our approach also integrates social and pragmatic factors. Beliefs are influenced by identity, social status, and interpersonal dynamics; fears of ostracism or admitting errors can increase resistance to change. Our hope is that through such modelling, we can take steps toward developing general frameworks that explain not just isolated sources of non-Bayesian belief updating, but also the inherent trade-offs between competing considerations associated with changing our minds.

\subsection{Contributions and Paper Structure}

Our primary contribution is the proposal of a variational cost functional for belief revision that models the influence of pragmatic affordances and cognitive costs on human belief updating. Secondly, we present results from simplified computational experiments demonstrating how varying conservatism and likelihood weighting parameters qualitatively exhibit phenomena such as confirmation bias, evidence search asymmetries, and attitude polarisation. Finally, we discuss the implications of our model and promising future directions.

\section{Related Literature}

\textit{``There is considerable evidence that people are more likely to arrive at conclusions that they want to arrive at, but their ability to do so is constrained by their ability to construct seemingly reasonable justifications for these conclusions.} (Kunda, 1990)~\cite{kunda1990case}.

\subsection{Decision-Theoretic Models of Belief Updating}

Drawing on frameworks of decision making, human belief revision is increasingly being treated as a value-based decision~\cite{kruglanski2020all,priniski2022bayesian,sharot2023Why}. On this view, beliefs are updated not purely based on their accuracy, but are associated with a \textit{utility}. The utility of holding a belief is derived from the outcomes it leads to, which can be internal (emotional comfort, positive feelings) or external (acceptance within a community, job opportunities)~\cite{sharot2023Why}. This is supported by several arguments highlighting the centrality of affect in decision-making and belief-updating~\cite{sharot2016forming,kiverstein2024desire,deane2024computational}. Certain beliefs may give rise to utility in proportion to how well they track or predict reality, in which case, there is an incentive for the agent to seek truthful beliefs. Other beliefs may demand that the agent confabulates an elaborate yet tenuous narrative that coincidentally supports their desired conclusion. In other words, a belief's usefulness can be orthogonal to its truthfulness.

\subsection{Cognitive Costs}

In addition to the pragmatic incentives that shape belief updating, there are unavoidable costs that any agent must pay to change their beliefs. These costs have been studied from the perspective of bounded/resource/computational rationality, where the presence of some form of cost associated with cognition is explicitly modelled in an agent's decision-making~\cite{simon1972theories,ortega2013thermodynamics,ortega2015information,lewis2014computational,gershman2015computational,lieder2020resource,zhu2023computation,parr2023cognitive}. Belief updating can be understood as a thermodynamic process involving transitions between mental states, where each transition incurs unavoidable dissipative costs~\cite{fields2024making}. These costs arise from fundamental physical principles governing information processing in biological systems at the level of neural computation and metabolic energy expenditure~\cite{wolpert2016free,wolpert2024stochastic}.

The transition between belief states involves both work-like and heat-like components. The work-like component corresponds to the directed shift of the belief state, while heat dissipation occurs in the form of entropy production during the transitions between belief states in finite amounts of time~\cite{andresen2011current}.
The total dissipation produced by belief updating can be quantified as the difference between the reversible work theoretically possible and the actual work captured during the transition process. This represents the unavoidable cost of finite-time belief changes~\cite{parrondo2015thermodynamics}.

This thermodynamic perspective helps to explain why, other considerations being equal, rapid belief changes tend to be more costly and inefficient compared to gradual updates. The system must balance the speed of belief revision against the increased dissipative costs of rapid change. This intuition can be made more precise using concepts from finite-time thermodynamics~\cite{andresen2011current}. The total entropy production in a sequence of step-equilibrations is bounded by $\Delta S^u \geq \frac{L^2}{2K}$, where $L$ is the thermodynamic length of the belief change pathway and $K$ is the number of intermediate equilibration steps~\cite{salamon2023more}. Thus, increasing the number of steps decreases the lower bound on total entropy production, permitting more efficient pathways of belief change. The brain appears to possess several remarkable features that aid in minimising these costs. For instance, the efficient coding hypothesis suggests that neural representations of sensory information are structured to minimise the number of neuronal spikes required to transmit a given signal~\cite{barlow1961possible}.

Understanding these fundamental thermodynamic constraints provides insight into why belief change can be so difficult even in the presence of contradictory evidence. The brain must carefully balance the energetic and informational costs of updating against the potential benefits. It is unclear precisely how significantly the thermodynamic costs of belief change contribute to this effect, and it would be worthwhile empirically investigating how such considerations can contribute to and explain belief inertia.

\subsection{Social Costs}

Human beliefs serve not only as internal models of the world but also as social signals and commitments. In active inference and variational learning frameworks, agents update beliefs to minimise surprise or prediction error, yet these updates occur in a social context where beliefs fulfil both epistemic (truth-seeking) and social-coordination functions \cite{mercier2017enigma, Bicchieri2014, Williams2021, albarracin2022epistemic, Constant2019, Veissiere2020, Vasil2020}. Believing (or disbelieving) certain propositions can grant individuals emotional comfort or group acceptance, independent of the belief’s accuracy \cite{sharot2023Why}. This dual role means that an agent’s posterior after observing new evidence is not determined by epistemic considerations alone, but also by the expected social and personal utilities associated with holding particular beliefs \cite{kunda1990case, sharot2023Why, Albarracin2024, GueninCarlut2024}. Consequently, standard Bayesian updates, which are focused purely on data and prior likelihood, are often tempered by an additional motive: to align with valued identities and norms that confer utility on the belief state \cite{mercier2017enigma, Williams2021, Constant2019, GueninCarlut2024}. This insight echoes the idea that \textit{all thinking is ``wishful'' thinking} to some extent, with motivational imperatives modulating inferential processes \cite{kruglanski2020all}. The free energy minimised during belief updating thus implicitly includes not just accuracy-related (surprisal) terms but also pragmatic terms capturing the work required to overcome cognitive inertia and social repercussions \cite{Bouizegarene2024, albarracin2022epistemic}.

Changing one’s mind can threaten group affiliations and invite real or perceived social sanctions (e.g., loss of status, trust, or membership) \cite{Bicchieri2014}. Beliefs often function as markers of group identity, so revising a key belief may signal disloyalty or value misalignment, incurring social costs like ostracism or ridicule. Anticipation of such costs creates a strong deterrent to belief revision, especially for identity-linked beliefs maintained by tight-knit communities and normative expectations \cite{mercier2017enigma, albarracin2022epistemic}. Indeed, social norms enforce a kind of epistemic conformity: individuals internalize the expectation that they ``ought'' to hold certain beliefs to remain in good standing \cite{Bicchieri2014, GueninCarlut2024}. From a decision-analytic perspective, the utility of a belief therefore includes not only its truth-tracking benefits but also its social payoff. A false or unfounded belief might persist if it brings social acceptance or emotional relief, whereas a truthful belief might be resisted if it carries stigma or existential dread. Accordingly, belief change in social contexts resembles a form of motivated reasoning: agents are inclined to arrive at the conclusions they want (or need) to reach, as long as they can justify them to themselves and others \cite{kunda1990case}. Here, ``wants'' are not arbitrary whims but structured by social identity and normative pressures—what one wants to believe is often what one’s group wants one to believe. An agent will unconsciously search for justifications to retain beliefs that serve valued social goals (e.g. solidarity, consistency, pride) and discount evidence that threatens those goals. The free energy landscape is warped by social potential energy. Certain directions of belief change appear steep (costly) due to the interpersonal consequences associated with them.

Empirical research supports these principles. For instance, people consistently overestimate the severity of the social sanctions they will face for changing a politically charged belief, leading to excessive self-censorship and public conformity \cite{Spelman2023}. In one set of their studies, U.S. partisans expected far more backlash from their in-group if they voiced a dissenting opinion than what actually materialised, with an average overestimation effect size of $d\approx0.87$. These inflated expectations of ostracism or punishment (sometimes stemming from an egocentric bias in social perspective-taking) make belief revision seem riskier than it truly is. Accordingly, individuals often stick to publicly defending their prior attitudes, even when privately grappling with contrary evidence. Social psychologists refer to this pattern as identity-protective cognition, wherein reasoning processes bend to protect the agent from the social identity costs of admitting error. The effect can become self-reinforcing. If everyone fears speaking up or changing their mind, the apparent unanimity of belief within the group remains unchallenged, further raising the perceived cost of dissent. Yet research also shows that these perceived social costs are malleable. Prompting individuals to reflect on their past loyalty and contributions to the group can reassure them that a change of mind will not irrevocably brand them as ``disloyal,'' thereby significantly reducing their concern about sanction and encouraging more open expression of revised beliefs. 

Beliefs are multi-functional cognitive tools that balance accuracy, utility, and inertia. They must at once represent the world (epistemic accuracy), support our emotional needs and moral values, and coordinate with our social milieu (utility), all while minimising drastic revisions that incur cognitive and social ``work'' (inertia). This perspective prepares us to interpret classic phenomena—confirmation bias, selective exposure to information, and attitude polarisation, not as inexplicable failures of rationality, but as strategic trade-offs given the agent’s objectives. An agent facing high costs for belief change will rationally exhibit a kind of conservatism. Agents will favour information that confirms existing beliefs and avoids provoking costly updates. Indeed, a confirmation bias in information-seeking can be seen as an adaptive strategy to preserve high-utility beliefs by selectively attending to congruent evidence and filtering out challenges. Experimental studies of selective exposure document that people spend more time with news and arguments that align with their preexisting attitudes than with those that contradict them, even when source credibility is controlled \cite{westerwick2017confirmation}. By skimming ``friendly'' evidence, individuals reduce the likelihood of encountering data that would demand painful social readjustments or internal value conflicts. Similarly, communities may become polarised when each side’s beliefs carry their own social rewards—members of opposing groups double down on group-consistent narratives, bolstering internal cohesion at the expense of cross-group accuracy. Over time, this self-reinforcing selection and interpretation of evidence drives group attitudes further apart, as each group lives in a bubble where maintaining their version of reality is pragmatically advantageous \cite{albarracin2022epistemic}. The following sections will explore how confirmation bias in evidence appraisal, asymmetrical information search, and polarisation dynamics emerge naturally once we acknowledge that changing one’s mind is not ``free.'' It incurs variational costs, paid in both cognitive effort and social capital, which a resource-rational mind navigates by carefully weighing when belief change is truly worth the price.

\subsection{Confirmation Bias}
Confirmation bias manifests through selective attention mechanisms, as shown in recent experimental work~\cite{prat2018selective,talluri2018confirmation}. Westerwick et al. demonstrated that when selecting political information online, participants spent more time with content matching their existing views, regardless of source quality~\cite{westerwick2017confirmation}. This bias emerged from participants' choices rather than the content itself. Building on this, \cite{talluri2018confirmation} revealed that making a categorical choice selectively enhanced sensitivity to subsequent evidence consistent with that choice, similar to attentional cueing effects. \cite{prat2018selective} proposed a neural mechanism for this bias, suggesting that choices direct feature-based attention to amplify processing of choice-consistent sensory evidence while suppressing inconsistent information. Together, these findings indicate that confirmation bias operates through early attentional selection rather than solely in later-stage decision processes.

\subsection{Motivated Reasoning}
Confirmation bias is a type of motivated reasoning, a process where information processing is biased toward achieving desired outcomes rather than accuracy alone \cite{kunda1990case}. Motivations can be accuracy-driven, encouraging unbiased reasoning, or directional, prompting strategies that reinforce existing beliefs, identity, or preferred conclusions. However, motivated reasoning remains constrained by plausibility; people select cognitive processes, such as memory retrieval and interpretation, that justify favoured conclusions rather than inventing implausible beliefs \cite{patterson2015motivated, ditto2009motivated, jain2000motivated}.

Individuals revise beliefs asymmetrically, giving more weight to confirmatory or emotionally favourable evidence than to equally informative negative evidence~\cite{little2025distinguish}. Motivated reasoners also selectively trust or avoid sources based on alignment with their views, effectively assigning lower reliability to disconfirming information. This acts like a biased Bayesian filter, reducing the impact of contradictory evidence on belief updating \cite{pilgrim2024confirmation}. 

Consequently, shared evidence can polarise rather than unify groups with opposing and even similar priors. When interpreting balanced evidence through a biased lens, individuals' initial beliefs often become more extreme, exacerbating attitude polarisation \cite{bartels2002beyond, albarracin2022epistemic}.

\subsection{Biased Reasoning}

Two recent frameworks have been proposed to explain some of the biases that occur in reasoning: coherence-based reasoning (CBR)~\cite{simon2025toward} and belief-consistent information-processing (BCIP)~\cite{oeberst2023toward}. Coherence-based reasoning posits that individuals strive to maintain a consistent and interconnected set of beliefs, minimising cognitive dissonance. More specifically, in CBR, a constraint-satisfaction network settles into an attractor by bidirectionally reshaping both beliefs and incoming information to maximise overall coherence. Crucially, strongly activated priors are harder to dislodge, so they often anchor the attractor state. Similarly, belief-consistent information processing describes the tendency to favour information that aligns with pre-existing beliefs, a process that is less cognitively demanding than evaluating and integrating contradictory evidence. BCIP is a special case of CBR’s coherence construction under conditions of dominant priors~\cite{oeberst2025belief}.

These two frameworks can be reconciled with our account through the lens of cognitive economy. Our variational cost framework formalises this principle by suggesting that altering one's beliefs incurs a cognitive cost, quantified by the KL divergence, which measures the informational distance between prior and posterior beliefs. Moreover, the weight of other firmly held beliefs can be explained by the presence of costs associated with revising more strongly held beliefs. For example, the expected cost associated with modifying a fundamental belief such as ``I make correct assessments of the world'' would be increased levels of doubt about the reliability of one's assessments, potentially leading to greater general levels of uncertainty and the accompanying negative affect that often arises. 

In this sense, both coherence-based reasoning and belief-consistent information processing can be viewed as cognitive strategies that minimise the costs that we describe. By maintaining coherence and selectively processing information, individuals reduce the ``informational work'' required to update their mental models of the world, thereby avoiding the significant pragmatic and cognitive expenditures associated with belief revision. In essence, these frameworks highlight different facets of the same underlying drive to manage cognitive resources efficiently, where the perceived utility of a belief is weighed against the inherent costs of mental reorganisation.

\section{A Motivated Variational Belief Change Model}
\label{sec:model}

As a starting point, we take inspiration from \textit{variational inference}~\cite{wainwright2008graphical,blei2017variational}, which underpins the mathematical formalism of the Free-Energy Principle and Active Inference~\cite{friston2010free,friston2023free}. In the standard Bayesian paradigm, the goal is to infer a posterior belief $p(s \mid o)$ about the state of the world $s$ given an observation $o$. According to Bayes' rule, finding this posterior requires one to compute the model evidence or marginal likelihood $p(o)$, which is an intractable problem in general. Variational inference aims to reformulate this problem by recasting it as an \textit{optimisation problem} over a variational family $\mathcal{Q}$ of probability distributions. The objective function of this optimisation problem is the negative \textit{evidence lower bound} (ELBO) or \textit{variational free energy} (VFE), and is given by
\begin{align} \label{eq:accuracy_complexity}
    F[q(s),o] &= - \underbrace{\mathbb{E}_{q(s)}[\log p(o \mid s)]}_{\text{Accuracy}}\ +\ \underbrace{\DKL{q(s)}{p(s)}}_{\text{Complexity}}\\ \label{eq:energy_entropy}
    &= \underbrace{-\mathbb{E}_{q(s)}[\log p(s,o)]}_{\text{Energy}}\ - \ \underbrace{H[q(s)]}_{\text{Entropy}}.    
\end{align}

The decomposition in Equation \ref{eq:accuracy_complexity} highlights a key tension between the expected log likelihood of the observation (accuracy) and the KL divergence from the prior to the variational posterior (complexity). In particular, this complexity acts as a \textit{regulariser} on the agent's posterior beliefs, penalising models that differ more from the prior. 

Core to the description of any agent is a description of its boundary, also commonly known as its \textit{Markov blanket}. The Markov blanket of an object describes the interface via which it is coupled to its environment. According to the Free Energy Principle (FEP), the internal paths of systems possessing a Markov blanket can be viewed as probabilistic beliefs about external paths, and the internal and active paths of the system appear to minimise its VFE~\cite{friston2023path}. When moving to descriptions of agents, however, the system's internal states embody not only a predictive model of the world, but also \textit{preferences} over possible configurations of the agent. This is where \textit{active inference} (AIF) comes into the picture.

\begin{figure}[t]
    \centering
    \includegraphics[width=0.8\linewidth]{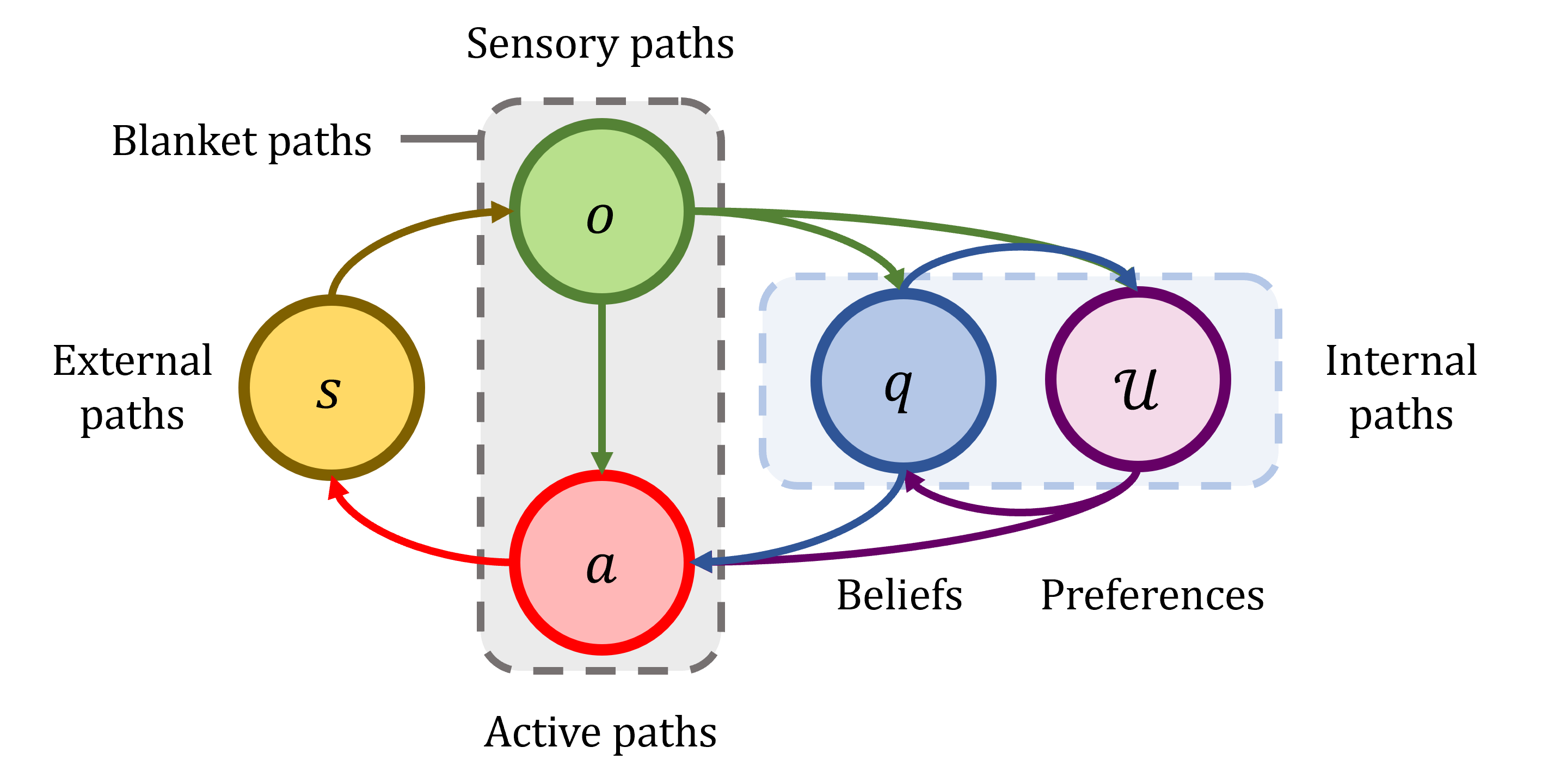}
    \caption{A depiction of the causal influences of different components of the agent-environment pair on each other. External paths $s$ represent states of the world external to a system or agent under consideration. This system possesses a Markov blanket, which separates internal from active paths and can itself be divided into sensory and active paths. We further assume that internal paths consist of two distinct components: beliefs and preferences, which mutually influence each other, are influenced by sensory paths, and in turn influence active paths.}
    \label{fig:agent_diagram}
\end{figure}

AIF extends the FEP to recognise the role of the \textit{actions} that agentic systems can perform to influence the environment and, vicariously, their observations~\cite{da2020active,parr2022active,da2024active}. Cast here in the variational perspective, the objective function that is posited to drive decision-making in AIF is the \textit{expected free energy} (EFE), which is a functional of a policy (sequence of actions) $\pi$, and is given by

\begin{equation}\label{eq:EFE}
    G(\pi) = -\underbrace{\mathbb{E}_{q(s,o\mid \pi)}\left[\DKL{q(s\mid o,\pi)}{q(s\mid \pi)}\right]}_{\text{Epistemic value}} - \underbrace{\mathbb{E}_{q(o\mid \pi)}[\log \tilde p(o)]}_{\text{Pragmatic value}},
\end{equation}

where $\tilde p(o)$ is a probability distribution representing the agent's preferences over their own observations, commonly known as a \textit{prior preference}~\cite{da2020active}. Here, we propose to extend this picture by considering the implications of assuming that agents have \textit{preferences about their own beliefs}, and develop a mathematical framework for describing the ensuing implications for belief updating. In other words, we suggest that the $C$ matrix, which is used in the active inference literature to parameterise the preference prior~\cite{da2020active,heins2022pymdp,parr2022active}, can be extended to be defined over the agent's own beliefs as well, rather than only observations/states.

For the purposes of this study, we focus on the mechanisms that drive belief change in agents. In particular, we are interested in the mapping from sensory states to belief states. We investigate the consequences of assuming that this mapping is comprised of two key components. The first component is a preference satisfaction component, represented here by an "expected utility" term, which can be related to prior preferences through a softmax transformation~\cite{da2023reward}. The second component is a direct cost for belief updating, which is quantified by the KL divergence from the agent's prior beliefs to their posterior beliefs. 

We will further assume that agents' preferences are grounded only in particular paths, and not directly on external paths, which is in concordance with an affect-driven view on motivation~\cite{shenhav2024affective,sennesh2025affective}. Under these assumptions, an agent's preferences can be described mathematically by a \textit{utility functional} $\mathcal U\;:\;\mathcal{Q} \times \mathcal{O} \to \mathbb R$ defined over observations and \emph{beliefs}, but not external states.

\begin{figure}[t]
    \centering
    \includegraphics[width=0.8\linewidth]{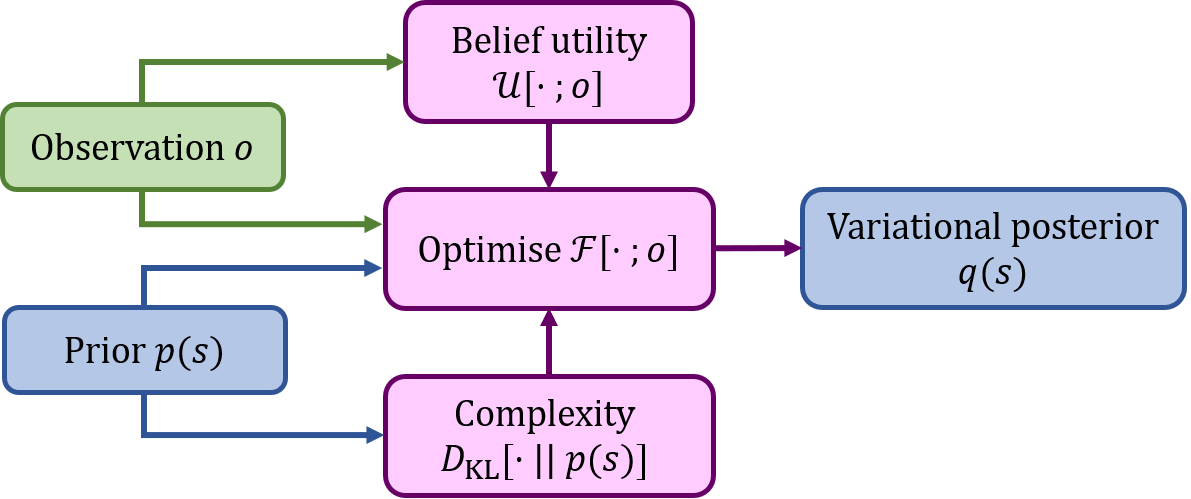}
    \caption{Schematic depicting the components involved in our proposed model of belief updating. Prior beliefs and observations act as inputs to a process that optimises a balance between a belief utility functional and a complexity term, measuring the KL divergence from prior to the ultimate posterior beliefs.}
    \label{fig:update_mechanism}
\end{figure}

Given this, we model an agent's belief updating processes as a variational optimisation process that maximises the following functional of beliefs and observations:
\begin{equation}
  \mathcal F[q(s),o]
  \;=\;
  \underbrace{\mathcal U[q(s),o]}_{\text{belief utility}}
  \;{-}\;
  \lambda\,\underbrace{\DKL{q(s)}{p(s)}}_{\text{complexity}},
  \label{eq:F_general}
\end{equation}
where $\lambda\geq 0$ is a parameter determining the relative strength of the cost of belief updating to the belief utility term. The belief utility term may or may not depend on the accuracy of the model, and depending on its form, can give rise to different belief updating behaviours. Under this model, taking $\lambda \to 0$ induces a belief update that is purely driven by belief utility, which is akin to assuming that the agent is able to instantaneously and effortlessly convince themselves of whatever they wish to believe. On the other hand, taking $\lambda \to \infty$ increases the cost of updating to the point that the agent is no longer able to change their mind, under any circumstances.

Importantly, one particular form for the belief utility that we investigate is a linear combination of what we term an \textit{affective utility} and a weighted expected log-likelihood or \textit{accuracy} term, which takes the following form:
\begin{equation} \label{eq:U_affective}
    \mathcal U[q(s),o] = \underbrace{U[q(s),o]}_{\text{affective utility}} + \alpha\ \underbrace{\mathbb{E}_{q(s)}[\log p(o|s)]}_{\text{accuracy}},
\end{equation}
where $\alpha \geq 0$ is a \textit{likelihood weighting} parameter, which determines the extent to which the agent's final belief distribution explains the data it has observed. A higher value of $\alpha$ can be interpreted as a stronger desire to arrive at beliefs that explain the observed data well. Moreover, for constant affective utility functions and $\alpha = \lambda =1$, we recover the VFE as a special case of ``accuracy-motivated'' belief updating. In the following section, we study the predictions made by adopting the belief utility functional given in Equation \ref{eq:U_affective}.

\section{Experiments and Results}

To study the implications of our proposed model on how motivated agents update their beliefs, we conducted a series of minimal experiments using categorical distributions\footnote{The code for generating the experimental results can be found at \url{https://github.com/dkhyland/motivated-variational-belief-updating}}. In all simulations, we consider how a single piece of evidence presented in the form of a likelihood distribution may be selected and subsequently influence the belief updating process. In particular, we demonstrate that under our model, several key features of human belief updating are qualitatively recovered. Moreover, our model can serve as a framework to generate testable predictions and simulations of human behaviour in various scenarios.

\subsection{How do different agents react to differing degrees of good vs bad news?}

\begin{figure}[t]
\begin{subfigure}{0.5\textwidth}
  \centering
  \includegraphics[width=\linewidth]{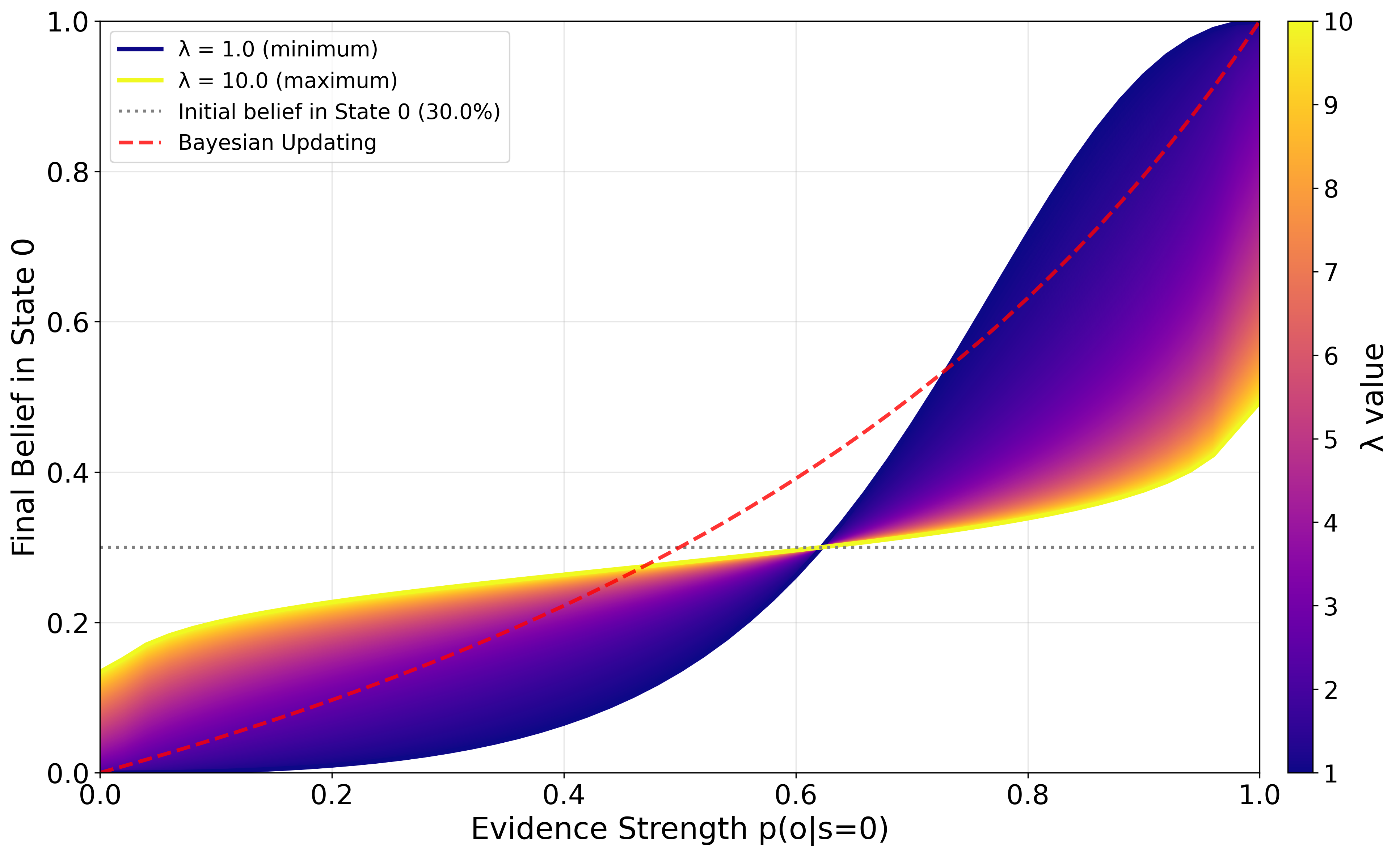}
  \label{fig:sfig1}
\end{subfigure}
\begin{subfigure}{0.5\textwidth}
  \centering
  \includegraphics[width=\linewidth]{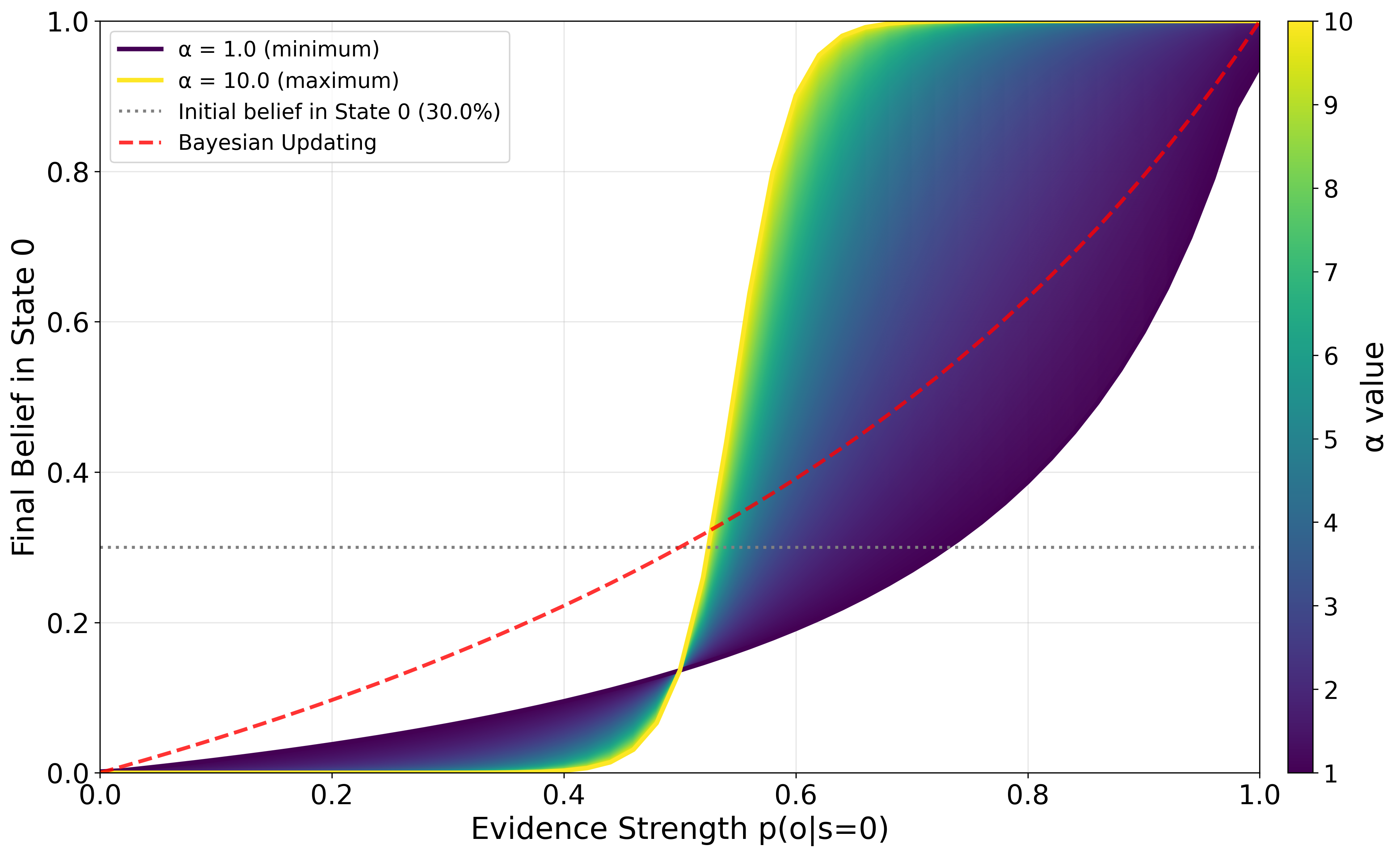}
  \label{fig:sfig2}
\end{subfigure}
\caption{Plots depicting how agents with different conservatism parameters $\lambda$ and likelihood weight parameters $\alpha$ respond to evidence that confirms or contradicts their belief preferences to varying degrees. Left: Final belief of the probability $q(s=0)$ of state 0 occurring as the evidence strength (in the form of a likelihood $p(o|s=0)$) varies from 0 to 1 for different values of $\lambda$. Right: Final belief $q(s=0)$ as evidence strength varies for different values of $\alpha$.}
\label{fig:motivated_update_plots}
\end{figure}

In our first set of experiments, we aim to understand and illustrate the effects of varying the strength of evidence, the conservatism parameter $\lambda$ and the likelihood weight parameter $\alpha$ on belief updating. In this scenario, an agent begins with an initial prior over the outcomes of a Bernoulli random variable (i.e., a biased coin flip) specified by $p(s = 0) =  0.3$, and receives evidence of varying strengths in the form of a likelihood $p(o \mid s)$. For Bernoulli random variables, we will assume that the hidden state $s$ may take the values $0$ or $1$. The agent updates their beliefs to minimise the objective in Equation \ref{eq:F_general} under the belief utility given in Equation \ref{eq:U_affective}. 

In Figure \ref{fig:motivated_update_plots}, we plot the final belief in state 0 as we vary the evidence strength $p(o \mid s=0)$ between 0 and 1 along the $x$ axis and the values of $\lambda$ (left) and $\alpha$ (right) as a spectrum for $\lambda \in [1,10]$ and $\alpha \in [1,10]$, along with the Bayesian update. From the left plot, we observe that higher values of $\lambda$ lead to updates that are closer to the prior, whereas lower values of $\lambda$ lead to updates that are more sensitive to the affective utility. From the right plot, the opposite effect is observed -- higher likelihood weights lead to more sensitivity to the evidence, and lower likelihood weights increase sensitivity to the affective utility.

\subsection{How do the relative strengths of belief utility and conservatism affect the selection of evidence?}\label{sec:evidence_selection}

In this study, we demonstrate the presence of a form of confirmation bias in our model, and seek to understand how different components of the model affect the selection of evidence in our motivated agent. In particular, recall that a crucial tenet within active inference is that agents are active sense-makers, selecting evidence to resolve uncertainty in both specific and non-specific manners in order to develop a better model of the world and achieve their ultimate objectives. In this experiment, we extend this notion to include the motivated selection of evidence to either confirm or disconfirm an agent's preferences.

\begin{figure}[t]
    \centering
    \includegraphics[width=\linewidth]{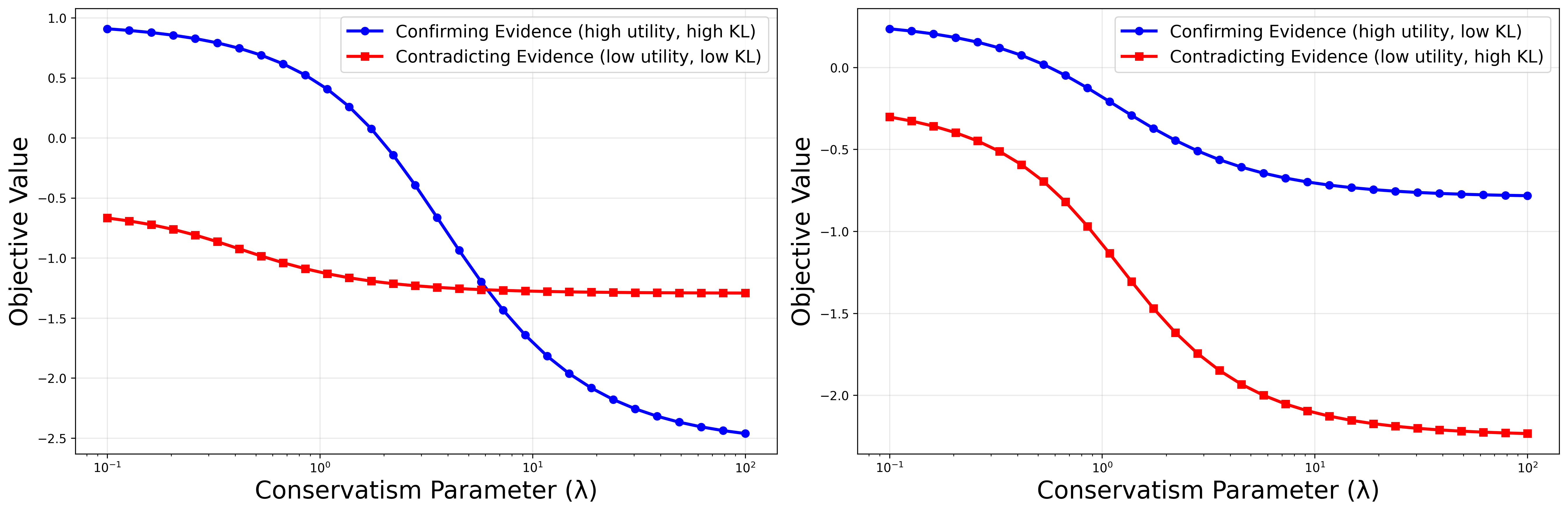}
    \caption{Plots of the objective value for Scenarios 1 and 2 with different combinations of evidence. In both scenarios, we fix $\alpha=2.0$ and use a linear affective utility functional with $U[q(s),o] = q(s=0)$. Left: Scenario 1, where Evidence A has high utility but high KL from the prior and Evidence B has low utility but low KL from the prior. Right: Scenario 2, where Evidence A has high utility and low KL, whereas Evidence B has low utility and high KL.}
    \label{fig:evidence_selection_threshold}
\end{figure}

We consider what happens when an agent with a linear affective utility who prefers to believe that $p(s=0)=1$ is presented with two pieces of evidence. We studied two different scenarios for what these pieces of evidence may be. In Scenario 1, the first piece of evidence (Evidence A) is \textit{`confirmatory'}, in the sense that it provides evidence for the agent's desired belief, but is further (induces updates with a larger KL divergence) from the agent's prior compared to the second piece of evidence (Evidence B). Evidence B is \textit{`contradictory'}, in the sense that it is evidence against the agent's desired belief but is closer to the agent's prior. In Scenario 2, Evidence A has both a higher affective utility and induces updates with a lower KL from the prior to the posterior. In Figure \ref{fig:evidence_selection_threshold}, the objective value is plotted as we sweep across values of $\lambda \in [0.1,100]$. In Scenario 1, we observe a threshold at which the agent switches from selecting confirmatory evidence to selecting contradictory evidence, whereas this does not occur in Scenario 2. Intuitively, this is because for low values of $\lambda$, the utility term dominates belief updating, but for higher values of $\lambda$, the cognitive cost term dominates. In contrast, when both the utility component is higher and cognitive costs are lower for one piece of evidence over the other, there is never a reason for the agent to choose to observe disconfirmatory evidence. 

\subsection{How do belief conservatism and likelihood weighting affect attitude polarisation?}

\begin{figure}
    \centering
    \includegraphics[width=\linewidth]{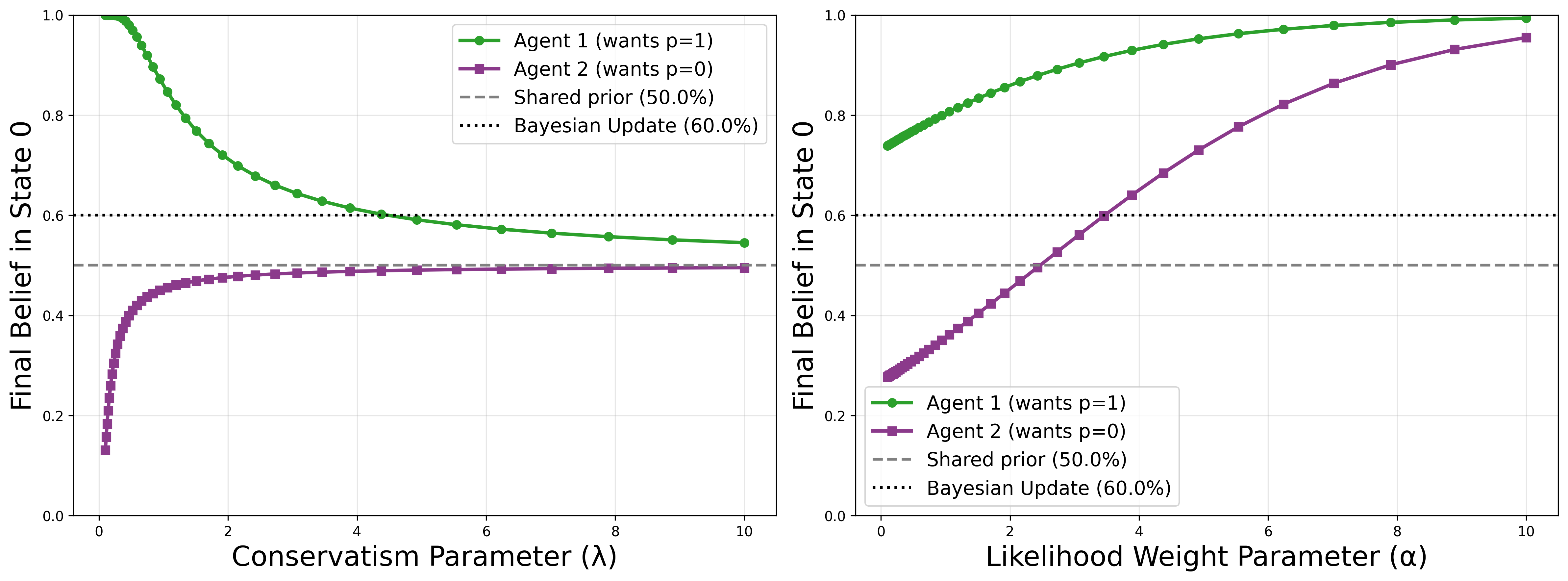}
    \caption{Plots of attitude polarisation effects between two agents who begin with the same prior beliefs and observe the same evidence, but have different affective utilities. Agent 1 linearly prefers to believe that $q(s=0)=1$, whereas Agent 2 linearly prefers to believe that $q(s=0)=0$. Left: Final beliefs as we vary $\lambda$ from 0 to 10. Right: Final beliefs as we vary $\alpha$ from 0 to 10.}
    \label{fig:attitude_polarization}
\end{figure}

In our final experiment, we simulated a basic attitude polarisation scenario, where two agents, Agents 1 and 2, began with the same prior belief about the probability $q(s=0)$, and observed the same evidence in the form of a likelihood. Both agents were endowed with linear affective utility functions, but Agent 1 had a preference for believing that $q(s=0)=1$ and Agent 2 had a preference for believing that $q(s=0)=0$. Plotting the final beliefs after updating according to our model as we varied $\lambda$ and $\alpha$ independently in Figure \ref{fig:attitude_polarization}, we observe that for low values of both parameters, the two agents' final beliefs differed significantly, demonstrating a basic form of attitude polarisation. However, as we increase the two parameters, the agents' final beliefs converged toward similar (though not necessarily Bayes rational) beliefs.

\section{Discussion}

Though much work needs to be done in empirically validating instantiations of the framework, our findings lend plausibility to the idea that realistic belief updating is subject to significant inertia and bias, driven largely by the constraints imposed on human agents by both internal cognitive limitations and external structures.

Realistic belief revision is rarely drastic, especially in the presence of cognitive costs for updating. From a cognitive standpoint, rapid updates necessitate more complex neural rewiring and a higher cognitive load, which can overwhelm limited cognitive resources. Agents would tend to avoid such costly leaps. Incremental steps across belief space reduce immediate costs but may also cumulatively result in lower total energetic and informational expenditure.

Under this view, we hypothesise that gradual transitions are typically more sustainable and preferable overall. Such considerations could explain why individuals are naturally inclined to resist abrupt changes in their belief systems despite potentially strong contradictory evidence, reinforcing conservative patterns of information integration.

\subsubsection{Strategies for effective belief updating}
Our basic model suggests several strategic insights for promoting more effective belief updating. Given the high cost of large leaps in belief space, strategies should prioritise incrementalism. This involves structuring information exposure in manageable segments that progressively lead individuals towards desired beliefs, thereby reducing the energetic, cognitive, and social resistance to dramatic changes. Social networks should be leveraged strategically: encouraging cross-cutting social ties and diversity in informational environments can reduce the perceived social risks associated with belief change.

\subsection{Future Directions}
In considering future avenues for research based on our current findings, several promising directions are worth exploring.

\subsubsection{Completing the Action-Perception Loop}
So far, our model has focused on the processes involved in the updating of beliefs, taking into account sensory evidence. Our preliminary data selection investigation takes this a step further by demonstrating how decisions about what data to observe can influence decision-making. However, further work is required to fully integrate motivation into the perception-action loop.

\subsubsection{Addition of temporal considerations}
So far, our model has not explicitly incorporated the temporal aspect of belief updating, which we believe to be significant in modelling the various costs that must be taken into consideration. Indeed, several works have posited a central role of \textit{rates of change} in free energy/prediction errors as crucial to understanding affect~\cite{joffily2013emotional,fernandez2021affective}. Extending the model to account for the role of time would allow a more detailed analysis of how belief trajectories could be optimised, rather than single updates.

\subsubsection{Extensions to group dynamics}
Further work could more explicitly incorporate group dynamics, particularly focusing on how social networks influence belief inertia and revision costs. Future work could explore the degree to which group identity and perceived social costs shape belief stability, potentially replicating frameworks similar to those presented by \cite{albarracin2022epistemic} on epistemic communities. By examining how belief updates propagate through structured networks and assessing how identity-protective reasoning reinforces certain belief states, we can quantify the inertia inherent within closely knit communities. Moreover, evaluating the relative weight of belief confidence levels and their susceptibility to drift could provide deeper insights into the dynamics of belief evolution in social contexts. Such extensions may also clarify how networked beliefs reinforce each other, creating feedback loops that stabilise misinformation.

\subsubsection{Further empirical validation}
Empirical validation remains essential for confirming and refining our theoretical propositions. Future empirical work will rigorously test model predictions using controlled laboratory experiments, field studies, and simulation analyses. For instance, quantifiable predictions derived from our framework—such as the relationship between KL divergence, belief revision speed, and associated cognitive or social costs—could be tested experimentally by monitoring physiological or neural responses during belief updating tasks. Longitudinal field studies examining belief trajectories within real-world social groups could also provide valuable validation, providing insights into how incremental versus rapid belief changes correlate with tangible social and cognitive outcomes.

\begin{credits}
\subsubsection{\ackname} The authors would like to thank Lancelot Da Costa and Tomáš Gavenčiak for helpful discussions and feedback.

\end{credits}

\bibliographystyle{acm}
\bibliography{refs}

\clearpage

\appendix
\section{Analytical Solution for Optimal Belief Updates in the Linear Affective Utility Case}
\label{app:linear_utility_update}

In this appendix, we derive the closed--form optimal posterior that minimises the
variational objective introduced in \ref{sec:model}. Throughout, let $S$ be a finite set of latent states $s\in S$, $q(s)$ the candidate posterior, and $p(s)$ the fixed prior. Observed data are denoted by $o$ with likelihood $p(o\mid s)$. The objective functional to be minimised is
\begin{equation}
  \label{eq:full_objective}
  \mathcal F[q(s),o] \;:=\; \underbrace{U[q(s),o]}_{\text{affective utility}}\; +\; \alpha\, \underbrace{\mathbb{E}_{q}\big[\log p(o\mid s)\big]}_{\text{accuracy}}
      \; -\; \lambda\, \underbrace{\DKL{q(s)}{p(s)}}_{\text{complexity}}.
\end{equation}
Here $U[q(s),o]$ represents the \emph{affective} utility of a belief state $q(s)$ and observation $o$, and
$\alpha,\lambda$ modulate respectively the weight assigned to the epistemic evidence and the inertia (or cost) of deviating from the prior.

For the case of linear affective utilities, we have
\begin{equation}
  \label{eq:linear_utility_def}
  U[q(s),o] \;=\; \sum_{s\in S} c_s \, q(s),
\end{equation}
with coefficients $c_s\in \reals$ capturing the valence of believing state $s$.

We maximise~\eqref{eq:full_objective} under the normalisation constraint $\sum_{s} q(s)=1$.  Introducing a Lagrange
multiplier $\eta\in \reals$ gives the augmented Lagrangian
\begin{align}
  \mathcal L(q,o,\eta) &=  \sum_{s\in S} c_s \, q(s) + \alpha\, \mathbb{E}_{q}[\log p(o\mid s)] - \lambda\, \DKL{q(s)}{p(s)} + \eta \Bigl( 1 - \!\sum_{s} q(s) \Bigr).
\end{align}
Stationarity with respect to each $q(s)$ yields
\begin{align}
  0 &= \frac{\partial \mathcal L}{\partial q(s)}
     = c_s \;+\; \alpha\, \log p(o\mid s)
       \;-\; \lambda\,\Bigl[1 + \log\frac{q(s)}{p(s)}\Bigr]
       \; -\; \eta.
  \label{eq:stationarity_raw}
\end{align}
Solving~\eqref{eq:stationarity_raw} for $q(s)$ and exponentiating, we obtain
\begin{align}
  q(s)
  &= p(s)
     \exp\!\left[ \frac{1}{\lambda}\Bigl( c_s + \alpha\, \log p(o\mid s) - \eta \Bigr) -1 \right] \\
  &\propto p(s)
     \exp\!\Bigl[ \tfrac{1}{\lambda}\bigl(c_s + \alpha\, \log p(o\mid s)\bigr)\Bigr].
  \label{eq:q_propto}
\end{align}
Normalising with the partition function
\begin{equation}
  Z(o) \;:=\; \sum_{s'\in S} p(s')\,
      \exp\Bigl[ \tfrac{1}{\lambda}\bigl(c_{s'} + \alpha\, \log p(o\mid s')\bigr) \Bigr],
\end{equation}
we arrive at the optimal variational posterior
\begin{equation}
  \label{eq:optimal_q_linear}
  q^{\star}(s)
  = \frac{p(s)
          \exp\bigl[ \lambda^{-1}\bigl(c_s + \alpha\,\log p(o\mid s)\bigr)\bigr]}
         {Z(o)}.
\end{equation}

\clearpage

\section{Additional Figures}

\begin{figure}
\begin{subfigure}{\textwidth}
  \centering
  \includegraphics[width=\linewidth]{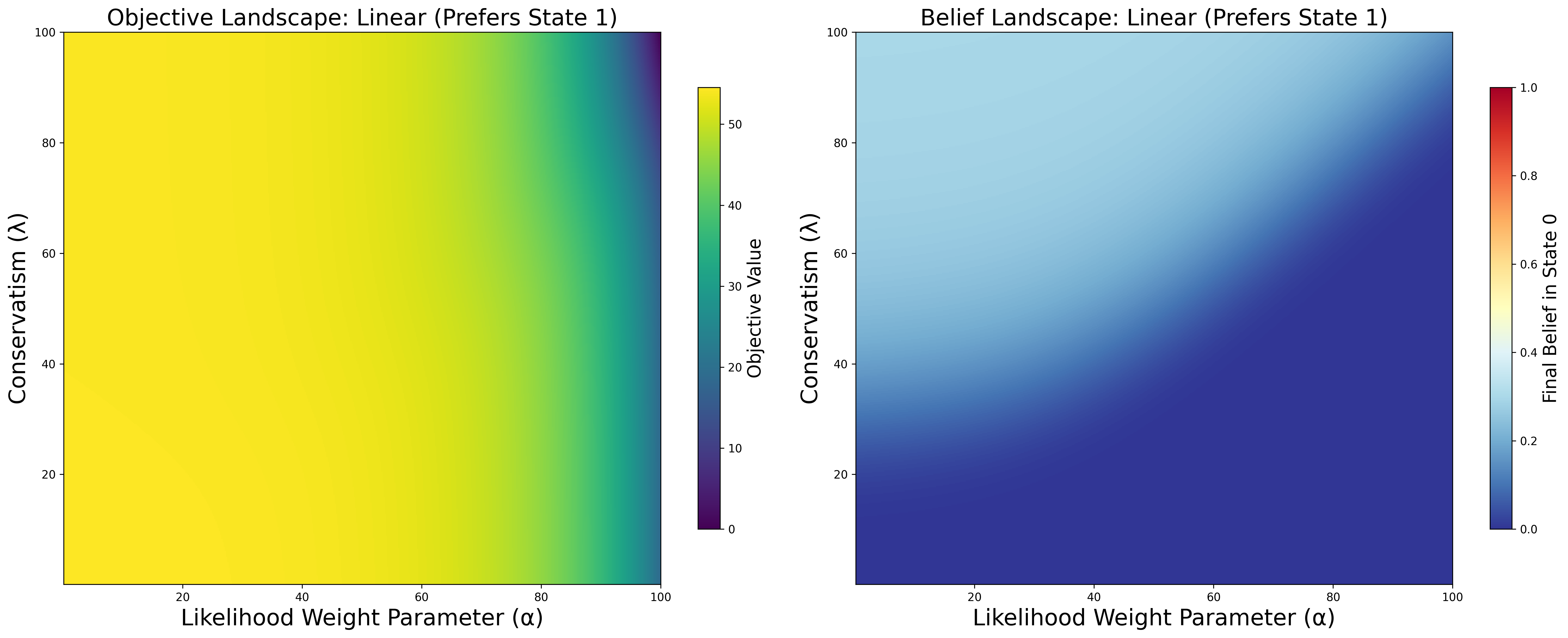}
  \label{fig:subfig1}
\end{subfigure}\\
\begin{subfigure}{\textwidth}
  \centering
  \includegraphics[width=\linewidth]{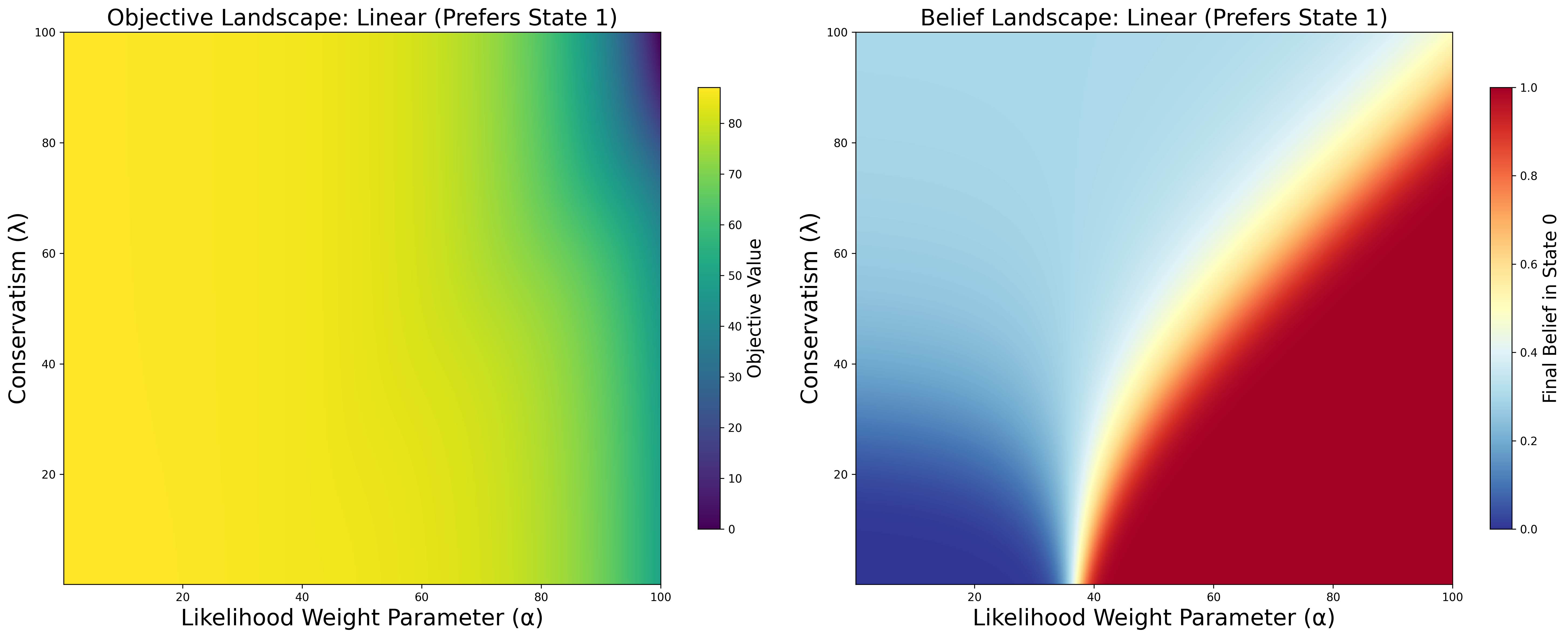}
  \label{fig:subfig2}
\end{subfigure}
\caption{Heatmaps depicting the variational objective and final belief landscapes for different $(\lambda,\alpha)$ pairs. The upper two panels depict the objective and belief landscapes (left and right, respectively) for evidence in the form of a likelihood where $p(o|s=0)=0.3$, and the bottom two represent the same but for evidence $p(o|s=0)=0.7$. For disconfirmatory evidence (top), }
\label{fig:prior_strength}
\end{figure}

\begin{figure}
    \centering
    \includegraphics[width=\linewidth]{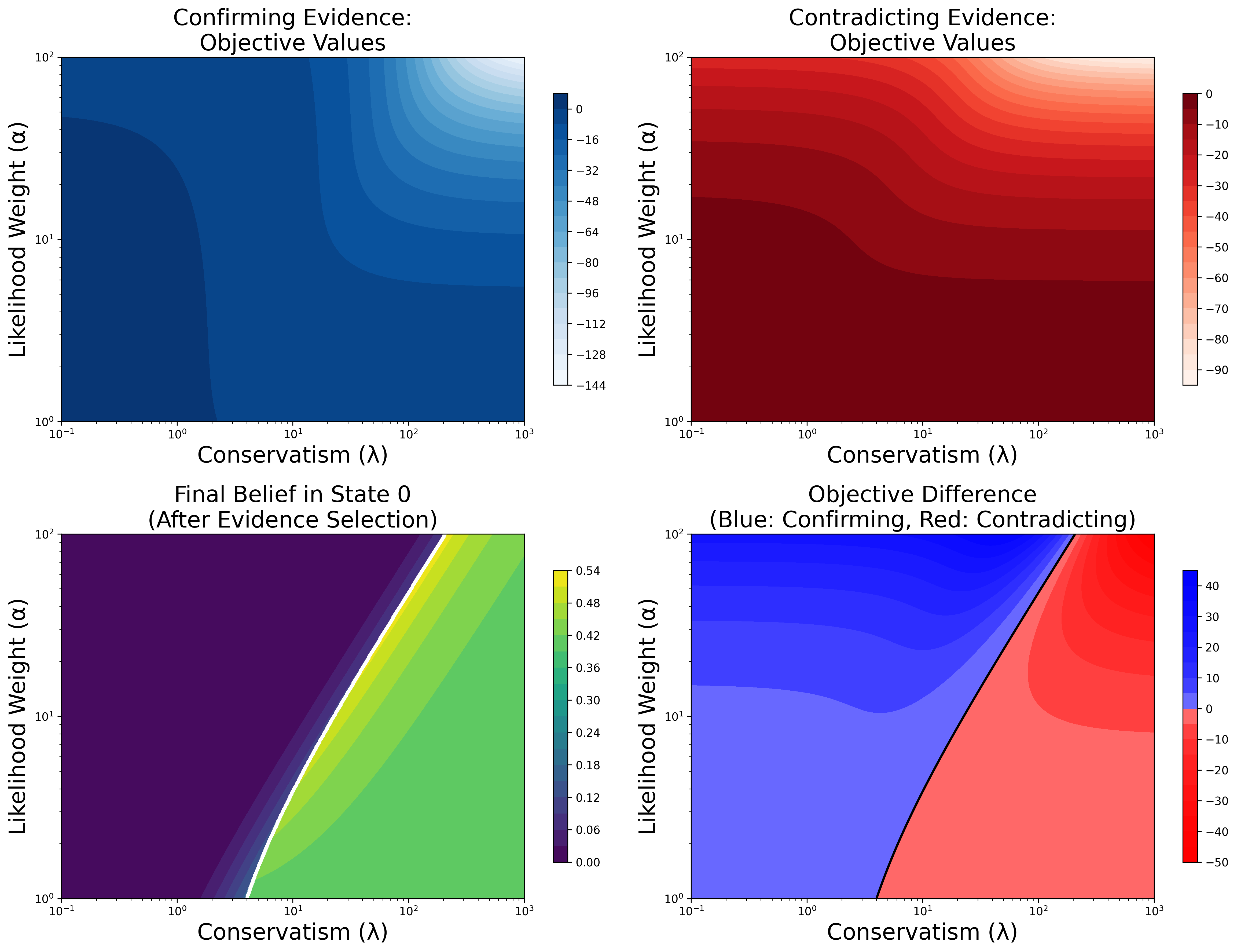}
    \caption{Contoured heatmaps depicting various quantities as we vary both the conservatism parameter $\lambda$ and the likelihood weight parameter $\alpha$ in Scenario 1 of wour second experiment (Section \ref{sec:evidence_selection}). Top left: heatmap of the objective landscape under confirmatory evidence, i.e., evidence that aligns with the agent's desired belief. Top right: heatmap of the objective landscape under contradictory evidence, i.e., evidence that contradicts the agent's desired belief. Bottom left: heatmap of the final belief $q(s=0)$ after evidence selection. The white line depicts the boundary at which the agent selects Evidence A (left of boundary) over Evidence B (right of boundary). Bottom right: heatmap showing the difference in the objectives for confirmatory and contradictory evidence. The black line again depicts the boundary between choosing Evidence A over Evidence B.}
    \label{fig:evidence_selection_heatmaps}
\end{figure}

\end{document}